\documentclass[letterpaper, 10 pt, conference]{ieeeconf}  
\usepackage{amsmath, amssymb} 

\usepackage{multirow} 
\usepackage{colortbl} 

\usepackage[table,xcdraw]{xcolor}

\usepackage{adjustbox}
\usepackage{array}
\usepackage{booktabs}

\usepackage[hyphens]{url}

\usepackage{rotating}
\usepackage{siunitx}

\usepackage{graphicx}

\usepackage{subcaption} 
\usepackage{subcaption}
\usepackage{array}
\usepackage{makecell}
\usepackage{adjustbox}

\usepackage{tabularx} 
\usepackage{balance}
\usepackage{xspace}
\usepackage{xcolor}
\usepackage{float}
\usepackage{xspace}
\usepackage{amssymb} 
\usepackage{pifont} 

\newcommand{\ourwork}{TrackOcc\xspace}
\newcommand{\ourtask}{Camera-based 4D Panoptic Occupancy Tracking\xspace}

\newcommand{\BN}{\mathbb N}
\newcommand{\SC}{{\cal C}}

\IEEEoverridecommandlockouts

\title{\LARGE \bf

TrackOcc: Camera-based 4D Panoptic Occupancy Tracking

}

\author{Zhuoguang Chen$^{1, 2, *}$, Kenan Li$^{2 ,*}$, Xiuyu Yang$^{2}$, Tao Jiang$^{2}$, Yiming Li$^{3}$, Hang Zhao$^{1,2,4, \dag}$
\thanks{*Zhuoguang Chen and Kenan Li contribute equally.}
\thanks{$^{\dag}$Corresponding author.}
\thanks{$^{1}$Shanghai Artificial Intelligence Laboratory, $^{2}$IIIS, Tsinghua University, $^{3}$New York University, $^{4}$Shanghai Qi Zhi Institute.}%
\thanks{Corresponding at: {\texttt{hangzhao@mail.tsinghua.edu.cn}}}
}

\begin{document}

\maketitle
\thispagestyle{empty}
\pagestyle{empty}

\begin{abstract}
Comprehensive and consistent dynamic scene understanding from camera input is essential for advanced autonomous systems. Traditional camera-based perception tasks like 3D object tracking and semantic occupancy prediction lack either spatial comprehensiveness or temporal consistency.
In this work, we introduce a brand-new task, Camera-based 4D Panoptic Occupancy Tracking, which simultaneously addresses panoptic occupancy segmentation and object tracking from camera-only input. Furthermore, we propose TrackOcc, a cutting-edge approach that processes image inputs in a streaming, end-to-end manner with 4D panoptic queries to address the proposed task. Leveraging the localization-aware loss, TrackOcc enhances the accuracy of 4D panoptic occupancy tracking without bells and whistles. Experimental results demonstrate that our method achieves state-of-the-art performance on the Waymo dataset. The source code will be released at \url{https://github.com/Tsinghua-MARS-Lab/TrackOcc}.
\end{abstract}

\section{INTRODUCTION}
A holistic and precise understanding of dynamic environments is essential for perception systems in robotics and autonomous vehicles.  A robust perception system needs to estimate the geometry, semantics, and identities of the current scene in a spatial-continuous and temporal-consistent way to interact with complex, changing 3D surroundings.

\begin{figure}[t]
  \centering

  \includegraphics[scale=0.43]{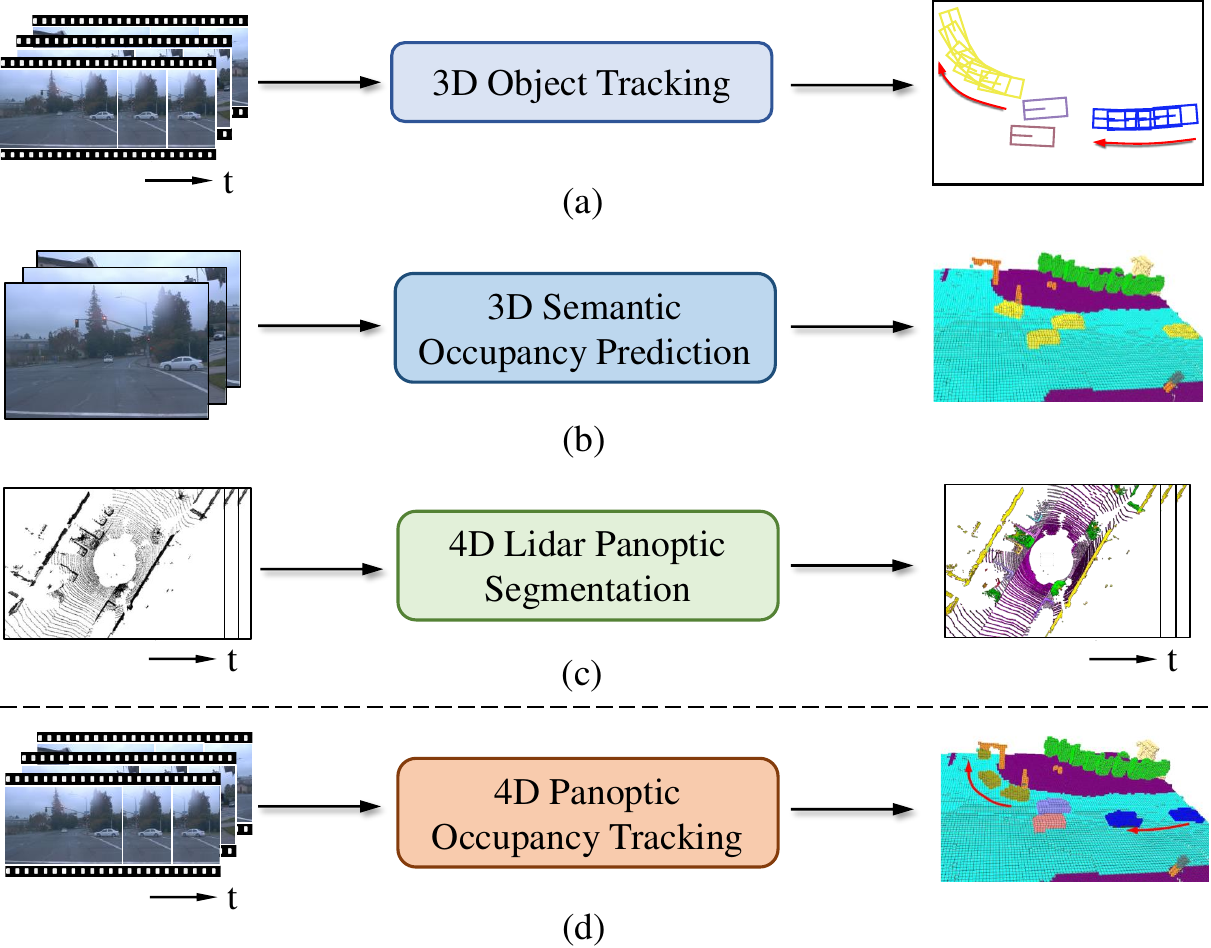}
    \caption{\textbf{Comparison of tasks for scene understanding.} (a) Outputs bounding box tracks from multi-view image sequences. (b) Predicts semantic labels for the occupancy volume from multi-view images. (c) Applies 3D panoptic segmentation on aggregated 4D LiDAR point volumes, followed by post-processing instance matching. (d) Our proposed task: Predicts temporally consistent panoptic labels of the occupancy from multi-view image sequences.}
    \label{figurelabel_fig1}
    \vspace{-4mm}
    \end{figure}

Previous efforts to achieve this goal include 3D object tracking~\cite{li2024DQTrack,zhangMUTR3DMulticameraTracking2022}, semantic occupancy prediction ~\cite{ma2023cotr,cao2022monoscene,huang2023tri,liu2023fully}, and 4D LiDAR segmentation ~\cite{Ayg2024DPLS, yilmaz24mask4former, kreuzberg2022stop}. As depicted in Fig.~\ref{figurelabel_fig1}, 3D object tracking (a) typically focuses on tracking objects across frames using bounding boxes. However, bounding boxes neglect fine-grained geometric details and struggle to represent general objects. On the other hand, occupancy-based tasks (b) offer a more complete representation of the 3D scene by incorporating both geometry and semantic information. Despite this advantage, existing approaches to occupancy prediction are mostly limited to spatial semantic perception, lacking a broader focus on panoramic temporal understanding.
4D Panoptic LiDAR Segmentation (4DPLS)~\cite{Ayg2024DPLS} tackles both semantic and instance segmentation in a 3D space over time, but it is constrained by the high cost of LiDARs and the limited sparsity of point clouds.

Different from previous efforts, we propose a brand-new task: \ourtask, as shown in Fig.\ref{figurelabel_fig1} (d). Unlike 4DPLS, which poses restricted semantic details, our task utilizes camera-only input, offering a cost-effective and easy-to-deploy solution that enhances spatiotemporal scene understanding. 

Furthermore, as shown in Fig.\ref{figurelabel_fig1} (c), previous methods for 4D LiDAR panoptic segmentation all conduct 4D LiDAR panoptic segmentation in a (3+1)D way, i.e., segment aggregated 3D volumes, and then post-process them to link the volumes~\cite{Ayg2024DPLS}. Instead, we introduce \textbf{\ourwork}, an end-to-end learning-based tracker approach for \ourtask. 

\ourwork processes image inputs in a streaming, end-to-end manner, eliminating the need for extensive post-processing. Specifically, we introduce 4D panoptic queries into \ourwork, enabling the prediction of occupancy with time-consistent panoptic labels in a unified framework. To ensure that the 4D panoptic queries capture spatially-accurate 3D features, we employ localization-aware loss to guide \ourwork towards the targeted areas, which significantly enhances the model's overall performance.

To summarize, our main contributions are three-fold:

\begin{itemize}
    \item To the best of our knowledge, we make the first attempt to explore a camera-based 4D panoptic occupancy tracking task, which jointly tackles occupancy panoptic segmentation and object tracking with camera input.
    \item We propose \ourwork, which uses \textit{4D panoptic queries} to perform the proposed task in a streaming, end-to-end manner. We also introduce a localization-aware loss to enhance the tracking performance.
    \item For fair evaluations, we propose the OccSTQ metric and build a set of baselines adapted from other domains. Experiments demonstrate that our \ourwork achieves state-of-the-art performance on Waymo dataset.

\end{itemize}

\section{RELATED WORK}

\subsection{3D Occupancy Prediction}
Occupancy representation, compared with 3D bounding boxes, offers finer geometric details and can assist in handling general, out-of-vocabulary objects~\cite{tian2023occ3d}.

The task requires to jointly estimate the occupancy state and semantic label of every voxel in the scene from images. Among the explorations, Occ3D~\cite{tian2023occ3d} offers a 3D occupancy dataset, followed by multiple solutions~\cite{ma2023cotr,liu2023fully, li2023fbocc}. Several recent works attempt to extend the task to panoptic prediction. SparseOcc~\cite{liu2023fully} utilized the inherent sparsity property of the scene to reduce the computational cost while offering an extension to panoptic occupancy prediction. PaSCo~\cite{cao2024pasco} extended the Semantic Scene Completion task to the panoptic segmentation. However, they cannot guarantee temporally consistent panoptic label prediction, specifically in tracking objects. Achieving this is a critical milestone for robots and autonomous vehicles, as it enables them to perceive and plan their paths effectively in the physical world.

\subsection{Video Panoptic Segmentation}
Video Panoptic Segmentation (VPS) is proposed in~\cite{kim2020VPS} to unify video semantic and video instance segmentation. In the beginning, numerous methods~\cite{kim2020VPS, voigtlaender2020siam} like VPSNet, depend on multiple sub-networks and complex post-processing (e.g. NMS, thing-stuff fusion). To alleviate these shortcomings, Recent research efforts for end-to-end video panoptic segmentation, such as~\cite{huang2022minvis, zhang2023dvis, ying2023ctvis, zhang2023dvis_plus} are extended from image panoptic segmentation method, incorporating specific trackers for this purpose. However, Since these methods rely on feeding video clips into the model, they can only handle short-term associations. Furthermore, these VPS methods are mainly applied to the 2D video domain and cannot be directly applied to the 3D world domain.

\subsection{3D Object Tracking}
3D Multi-object tracking (3D MOT) involves using multi-view images from cameras or point clouds from LiDAR to track multiple objects across frames. Building upon advances in 3D object detection~\cite{wang2022detr3d,liu2022petr,chen2023futr3d,lai2023spherical}, recent 3D trackers usually adopt the tracking-by-detection paradigm, associating trajectories with detected boxes through post-processing. However, this paradigm is heavily dependent on complex human designs and post-processing steps.
To overcome these limitations, MUTR3D~\cite{zhang2022mutr3d} has successfully extended the MOTR-type tracker to the 3D domain, achieving end-to-end 3D tracking with promising results. ADA-Track~\cite{ding2024ada} and DQTrack~\cite{li2024DQTrack} utilize query-based frameworks and learnable associations to detect and track objects in multi-view images. 

However, these methods are based on bounding boxes rather than occupancy, neglecting the finer geometric details and general objects. And they only focus on tracking \textit{thing} classes, ignoring recognizing \textit{stuff} classes, where we unify them in one framework.

\subsection{4D Panoptic LiDAR Segmentation}
The spatiotemporal interpretation in LiDAR point-based tasks has been extensively explored~\cite{yilmaz24mask4former, kreuzberg2022stop, marcuzzi2023ral-meem, zhu20234d}. Since the task was first defined. 4D-PLS~\cite{Ayg2024DPLS} initially proposed applying 3D panoptic segmentation to a frame of multi-scan-aggregated point clouds, aiming to reduce the need for explicit data association. Subsequently, methods such as Mask4Former~\cite{yilmaz24mask4former} have emerged, focusing on spatially compact instance segmentation with an additional 6-DOF bounding box regression. However, these approaches still rely on 3D segmentation, which cannot eliminate the need for offline post-association. As a result, it does not fully address the 4D panoptic segmentation problem and is not suitable for real-time streaming applications, limiting its use in mobile robots and autonomous driving systems.

   \begin{figure*}[t]
      \centering
      \includegraphics[scale=0.73]{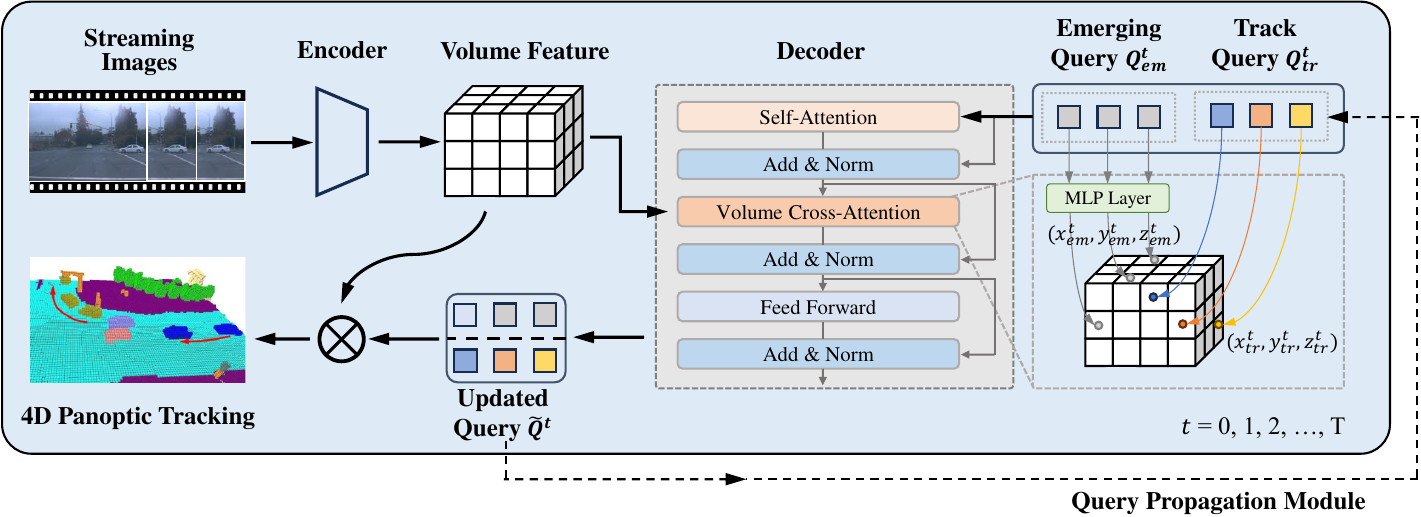}
      \caption{\textbf{Overall pipeline of \ourwork.} At each timestep, the encoder extracts multi-view image features and transforms them into 3D volume features. These volume features serve as critical context for updating the 4D panoptic queries via a designed decoder. The query propagation module facilitates efficient object tracking in a streaming, end-to-end manner. The symbol $\bigotimes$ denotes matrix multiplication.}
      \label{figurelabel_OverallArch}
   \end{figure*}

\section{METHOD}
\subsection{Problem Definition}
In this paper, we propose the \textbf{\ourtask} task, which jointly addresses occupancy panoptic segmentation and object tracking in both spatial and temporal domains. Specifically, the model takes as input a sequence of RGB images, denoted by \( \mathbf{I} = \{ \mathbf{I}_t^i, \mathbf{I}_{t-1}^i, \ldots, \mathbf{I}_{t-L}^i \}_{i=1}^{N} \), where \( N \) represents the number of surrounding cameras and \( L \) denotes the sequence length.
Given a predetermined set of \( C \) semantic classes, encoded as $\SC := \{0, \ldots, C-1 \}$, the task requires a neural network to map each voxel \( i \) in the grid to a pair $(c_i, z_i) \in \SC \times \BN$, where \( c_i \) represents the semantic class of voxel \( i \) and \( z_i \) denotes its instance ID. The semantic label set comprises both \emph{stuff} and \emph{thing} categories. When a voxel is labeled as \emph{stuff}, its corresponding instance ID \( z_i \) is irrelevant. 
Voxels in the free space are assigned a special \emph{free} stuff label.
The instance IDs \( z_i \) group voxels of the same class into distinct segments, which should persist throughout the entire sequence. The grid dimensions are \( X \times Y \times Z \), representing the height, width, and depth of the grid, respectively.

\subsection{Overall Architecture} \label{subsec:overallarch}

The overall proposed \ourwork architecture is depicted in Fig.~\ref{figurelabel_OverallArch} and comprises three main components. First, an encoder (Sec.~\ref{Encoder}) extracts volume features from multi-view images. Second, a novel query design facilitates both panoptic segmentation and instance tracking. Finally, a decoder (Sec.~\ref{Decoder}) interacts with the volume features and queries to generate the final 4D prediction in an end-to-end, streaming process. 

\subsection{Encoder} \label{Encoder}
The encoder constructs the 3D volume features from multi-view images by integrating an image feature extractor with a 2D-to-3D transformation module. The image feature extractor captures multi-scale features from multi-view images, which are then transformed into 3D volume representations $\mathbf{F} \in \mathbb{R}^{X \times Y \times Z \times D}$, where $D$ denotes the feature dimension per voxel. Extensive research has been dedicated to methods for transforming 2D image space into 3D volume space, including LSS~\cite{LSS}, BevDepth~\cite{li2023bevdepth}, BEVFormer~\cite{li2022bevformer}, FB-Occ~\cite{li2023fbocc}, and COTR~\cite{ma2023cotr}. In our model, we utilize COTR's encoder~\cite{ma2023cotr} as our 2D-to-3D transformation module. Importantly, \ourwork is flexible and compatible with all of these approaches.

\subsection{4D Panoptic Query} \label{Query Definition}

When designing queries for 4D Panoptic Occupancy Tracking, two main challenges arise:1) how to define queries for panoptic segmentation and instance tracking. 2) how to manage newborn and terminated instances.

\noindent\textbf{Query Definition.} We define two types of queries: emerging queries $Q_{\text{em}}$ and track queries $Q_{\text{tr}}$. Emerging queries are utilized for both \emph{stuff} and detecting newborn instances, while track queries are responsible for predicting all existing tracked instances. For simplicity, \emph{stuff} is assigned to emerging queries since it does not require tracking. The set of track queries is updated dynamically, and its size varies over time.
In the first frame, there are no track queries; only the fixed-length, learnable emerging queries are fed into the decoder. For subsequent frames, we input $Q^{\text{t}}$, which is the concatenation of track queries $Q_{\text{tr}}^{\text{t}}$ and the learnable emerging queries $Q_{\text{em}}^{\text{t}}$ into the decoder. These queries interact with the volume features in the decoder to generate updated queries 
\(\tilde{Q}^{\text{t}}=\tilde{Q}_{\text{em}}^{\text{t}} \cup \tilde{Q}_{\text{tr}}^{\text{t}} \) which are used to produce the final 4D predictions. The updated queries $\tilde{Q}^{\text{t}}$ are also passed to the Query Propagation Module to generate the track queries $Q_{\text{tr}}^{\text{t+1}}$ for the next frame.

\noindent\textbf{Query Propagation.} To handle instances that may appear or disappear in intermediate frames, we introduce the Query Propagation Module. This module manages the propagation of queries for newborn and terminated instances within our method. During training, for queries in $\tilde{Q}_{\text{em}}^{\text{t}}$, only queries matched to ground truth \emph{things} are converted into track queries, whereas those corresponding to \emph{stuff} are excluded. Queries in $\tilde{Q}_{\text{tr}}^{\text{t}}$ are removed if they have been matched to $\varnothing$ for $T_f$ successive frames.

During inference, we use the predicted classification scores to determine appearance of newborn instances and disappearance of tracked instances. For queries in $\tilde{Q}_{\text{em}}^{\text{t}}$, predictions with classification scores higher than the entrance threshold $\tau_{\text{1}}$ and corresponding to \emph{things} are retained. Queries corresponding to \emph{stuff} are discarded even if their classification scores exceed $\tau_{\text{1}}$. For queries in $\tilde{Q}_{\text{tr}}^{\text{t}}$, predictions with classification scores lower than the exit threshold $\tau_{\text{2}}$ for consecutive $T_f$ frames are removed.

\subsection{Decoder} \label{Decoder}
We detail the decoding process as follows. First, the queries interact with each other to enhance features using a self-attention~\cite{vaswani2017attention} layer. Then, they look up and aggregate volume features via a Volume Cross-Attention (VCA) layer. Due to the large input scale of the volume features, the computational cost of vanilla attention~\cite{vaswani2017attention} is high. Therefore, we adopt a VCA layer based on deformable attention~\cite{deformttn}, a resource-efficient attention mechanism where each query interacts only with its regions of interest in the volume.

Specifically, each emerging query $q \in Q_{em}$ predicts a 3D reference point $p \in \mathbb{R}^3$ using an MLP layer while each track query $q\in Q_{tr}$ maintains its corresponding reference point over time, and the volume features $F$ around these reference points are sampled. A weighted sum of the sampled features is then performed as the output. The process of volume cross-attention (VCA) can be formulated as:
\begin{equation}
\text{VCA}(q, F) = \sum_{j=1}^{J} w_j \cdot F(p + \Delta p_{j}),
\label{eq:vca}
\end{equation}
where $j$ indexes the sample point, and $w_j \in \mathbb{R}^{1}$ and $\Delta p_j \in \mathbb{R}^{3}$ are learnable weights and offsets.

Finally, a linear classifier, followed by a sigmoid activation function, is applied on top of the query embeddings to yield class probability predictions. Notably, the classifier predicts an additional ``no object'' category $\varnothing$ in cases where the query does not correspond to any region. For mask prediction, we obtain each binary volume mask prediction $m_i \in [0, 1]^{X \times Y \times Z}$ via a matrix multiplication between each query and the volume feature, followed by a sigmoid activation. The binary volume mask, combined with the class prediction, generates the panoptic occupancy prediction. For \emph{things}, the 4D predictions from the same track query across frames share the same ID.

\subsection{Loss function}

The optimization objectives of our approach focus on two main aspects: mask prediction and position prediction. We calculate the mask classification loss $\mathcal{L}_{mask-cls}$ for masks, and different types of queries are matched with the ground truth using different strategies. Track queries persist throughout the entire occupancy flow, enabling the assignment of ground truth to each target only once across all timesteps. Once created by inheriting from emerging queries, track queries are tied to the corresponding ground truth and remain unchanged.

Emerging queries, which transition between different timestamps, do not require explicit assignment at each timestep. Following Mask2Former~\cite{cheng2021mask2former}, we establish a correspondence between the ground truth and the prediction via bipartite matching, solved by the Hungarian algorithm~\cite{kuhn1955hungarian}.  Once the matching is established, we compute the mask classification loss for queries, denoted as $\mathcal{L}_{mask-cls}$, which includes the binary mask loss $\mathcal{L}_{mask}$ and the multi-class cross-entropy loss $\mathcal{L}_{cls}$.

\textbf{Localization-aware Loss.} Localizing and tracking each object accurately is both crucial and challenging. Previous works that rely on mask classification for 3D panoptic prediction, such as SparseOcc~\cite{liu2023fully}, have limited localization capabilities. Thus, we propose the localization-aware loss: 
\begin{equation}
\mathcal{L}_{loc} = \text{L1}(P', P),
\end{equation}
to improve the accuracy of query-specific 3D positions, since they are used in our VCA module to serve as the reference points for the attention. Note that the predicted points $P'$, which are regressed through an MLP, only come from the queries that are tied to \textit{thing} classes. We can obtain $P$ by easily calculate the centroid of each object. We observe that this enhancement significantly improves the metrics, as demonstrated in our experiments(Sec.~\ref{EXPERIMENTS}).

Overall, the final training loss is: 
\begin{equation}
\mathcal{L} = \mathcal{L}_{mask-cls} + \mathcal{L}_{loc}.
\end{equation}

\begin{table}[b]
\vspace{-4mm}
\caption{\textbf{Performance Comparison Between SparseOcc and COTR on 3D Panoptic Prediction.} Under the revision metric PQ$^\ast$, COTR outperforms SparseOcc in both small objects and stuff regions. C.C represents construction cone.}  
\label{table_3DPSCMetricCompare}
\centering
\small
\setlength{\tabcolsep}{5pt}
\renewcommand{\arraystretch}{1.4}
\centering 
\begin{adjustbox}{max width=0.95\columnwidth} 
\begin{tabular}{c|c|c|*{4}{c}} 
\toprule[1.25pt]
\multicolumn{1}{c|}{Methods}
& \multicolumn{1}{c|}{Metrics}
& \multicolumn{1}{c|}{Overall}
& \multicolumn{1}{c}{vehicle} 
& \multicolumn{1}{c}{pole} 
& \multicolumn{1}{c}{C.C.}
& \multicolumn{1}{c}{road} \\ 
\hline
\multirow{2}{*}{SparseOcc~\cite{liu2023fully}} 
& PQ & 5.0 & 12.4 & \textbf{0.1} & \textbf{2.6} & 13.6 \\
\cline{2-7} 
& ~PQ$^\ast$ & 12.8 & 21.6 & \textit{7.0} & \textit{13.8} & 34.4  \\
\cline{1-7}
\multirow{2}{*}{COTR~\cite{ma2023cotr}} 
& PQ & 13.6 & 13.7 & \textbf{0.5} & \textbf{0.6} & 76.9 \\
\cline{2-7} 
& ~PQ$^\ast$ & 22.4 &24.3 & \textit{15.0} & \textit{12.0} & 78.2 \\
\bottomrule[1.25pt]
\end{tabular}
\end{adjustbox}
\end{table}

\begin{table*}[t]
\caption{\textbf{\ourtask Performance on the Occ3D-Waymo Dataset~\cite{tian2023occ3d}.} MinVIS and CTVIS are query-based association methods, while AB3DMOT refers to the tracking-by-detection method. 4D-LCA denotes 4D LiDAR cross-volume association method in \cite{Ayg2024DPLS}. E2E represents end-to-end. Due to space limitations, only representative categories are selected. GO denotes General Objects.}
\label{table_4DCPOccBaselines}
\normalsize
\setlength{\tabcolsep}{3pt}
\renewcommand{\arraystretch}{1.3}
\centering
\begin{adjustbox}{max width=\textwidth}
\begin{tabular}{@{}c|c|c|c|c c c|c|c c c c@{}}
\toprule[1.25pt]
\multirow{2}{*}{Method} & \multirow{2}{*}{E2E} & \multirow{2}{*}{\textit{OccSTQ}} & \multicolumn{4}{c|}{OccAQ} & \multicolumn{4}{c}{OccSQ} & \\ \cline{4-12}
& & & Overall & Vehicle & Pedestrian & Cyclist & Overall & Sign & Building & Vegetation & GO \\ \hline
MinVIS~\cite{huang2022minvis} & \ding{51} & 9.8 & 3.3 & 4.0 & 1.5 & 3.1 & 29.1 & 10.6 & 39.1 & 34.0 & \textbf{11.7} \\
CTVIS~\cite{ying2023ctvis} & \ding{51} & 10.7 & 4.3 & 5.3 & 2.1 & 1.6 & 26.5 & 13.2 & 36.9 & 30.9 & 8.1 \\
AB3DMOT~\cite{Weng2020_AB3DMOT} & \ding{55} & 12.4 & 5.3 & 6.8 & 1.7 & 2.9 & 29.1 & 10.7 & 39.2 & 34.0 & \textbf{11.7} \\
4D-LCA~\cite{Ayg2024DPLS} & \ding{55} & 16.2 & 9.0 & 11.4 & \textbf{3.4} & 4.1 & 29.1 & 10.7 & 39.2 & 34.0 & \textbf{11.7} \\
\ourwork(ours) & \ding{51} & \textbf{20.0} & \textbf{13.5} & \textbf{18.1} & 3.1 & \textbf{4.6} & \textbf{29.4} & \textbf{13.6} & \textbf{43.2} & \textbf{40.0} & 9.2 \\
\bottomrule[1.25pt]
\end{tabular}
\end{adjustbox}
\end{table*}

\begin{table}[t]  
\caption{\textbf{Ablation Studies.} Comparison of results with different settings including Localization-aware Loss and the number of training frames.}  
\label{table_AblationStudies}  
\normalsize  
\setlength{\tabcolsep}{8pt}  
\renewcommand{\arraystretch}{1.1}  
\centering 
\begin{adjustbox}{max width=1\columnwidth} 
\begin{tabular}{cc|ccc}  
\toprule[1.25pt]  
w/ Loc. Loss & \#frames & OccSTQ & OccSQ & OccAQ \\ \hline  
\ding{55} & 3 & 16.5   & 28.9   & 9.4    \\ 
\ding{51} & 3 & 20.0   & 29.4   & 13.5   \\ 
\ding{51} & 5 &\textbf{20.6}   & \textbf{29.5}   &\textbf{14.4}   \\  
\bottomrule[1.25pt]  
\end{tabular}  
\end{adjustbox}
\vspace{-4mm}
\end{table}

\section{EXPERIMENTS} \label{EXPERIMENTS}
\subsection{Dataset}
We conduct our experiments on the Occ3D-Waymo~\cite{tian2023occ3d} dataset at 2 Hz, which provides semantic labels for 3D occupancy grids. This dataset includes 798 training scenes and 202 validation scenes, with a spatial range of $[-40m, 40m]$ for the $x$ and $y$ axes, and $[-1m, 5.4m]$ for the $z$ axis. The voxel grid size is $(0.4m, 0.4m, 0.4m)$, resulting in a resolution of $(200 \times 200 \times 16)$ for $(X, Y, Z)$. However, the dataset lacks panoptic labels, offering only semantic labels.

To generate 4D panoptic labels for our proposed 4D panoptic occupancy tracking task, we utilize the annotations from the Waymo Open Dataset~\cite{WODsun_2020_CVPR}. These annotations provide exhaustive 3D bounding boxes for vehicles, pedestrians, and cyclists, each with a unique tracking ID. We classify voxels corresponding to vehicles, pedestrians, and cyclists as \textit{thing} while all other categories are labeled as \textit{stuff}. Furthermore, \textit{thing} voxels are divided into instances, with voxels within a single 3D box grouped as one instance. For \textit{thing} voxels not located inside any 3D box, we assign them to the nearest 3D box. Finally, each voxel is assigned the tracking ID from its corresponding 3D box.

\subsection{Implement Details}
We adopt ResNet-50~\cite{He_2016_CVPR} as the image backbone with an input resolution of 256 × 704. Both self-attention and volume cross-attention is single-layer. The number of sampling points per reference point, $J$, is set to 4 for VCA in the Decoder. We put 200 emerging queries into each frame and the query dimension is consistently 256, both before and after the update. The query lifecycle is controlled by $\tau_{1}=0.3$, $\tau_{2}=0.25$, and $T_f=3$. All our experiments are conducted after training for 24 epochs with a learning rate of $2\times 10^{-4}$, using the AdamW~\cite{loshchilov2019decoupledweightdecayregularization} optimizer and a batch size of 1, on 8 NVIDIA GeForce RTX 3090 GPUs.

\subsection{Evaluation Metrics}
To evaluate the methods for \ourtask, and showcase the segmentation and tracking ability simultaneously, we propose the Occupancy Segmentation and Tracking Quality (OccSTQ) metric. Inspired by~\cite{Ayg2024DPLS, hurtado2020mopt, weber2021step}, the OccSTQ is derived from the STQ~\cite{weber2021step}, which is designed for video panoptic segmentation, effectively disentangles association and classification errors, and is well-suited for long sequences. Therefore, we adapt it as OccSTQ for our task.

We define a sequence of occupancy frames as \( \Omega = \{(\mathbf{v}, t) \mid \mathbf{v} \in X \times Y \times Z, t \in [0, T)\} \), where \( \mathbf{v} \) represents a voxel in a 3D spatial domain, and \( t \) denotes the timestep over a total duration. The ground truth is represented as \( gt(\mathbf{v}, t) \), and the prediction as \( pr(\mathbf{v}, t) \). OccSTQ consists of Occupancy Segmentation Quality (OccSQ) and Occupancy Association Quality (OccAQ), which measure segmentation and association quality respectively.

OccSQ is defined as mean IoU (Intersection-over-Union), similar to~\cite{weber2021step} and~\cite{Ayg2024DPLS}, but with voxels as the basic elements.

For OccAQ, we define the prediction for a specific identity in \textit{thing} class as:
$
pr_{id}(id) = \{(\textbf{v}, t) \mid pr(\textbf{v}, t) = (c, id), c \in C_{th}\},
$
and the ground-truth is defined analogously as $gt_{id}(id)$.
Thus the true positive associations (TPA) is:
$
\textit{TPA}(p, g) = \left| pr_{id}(id^{\prime}) \cap gt_{id}(id) \right|
$. It is important to note that the predicted $id^{\prime}$ and the ground-truth $id$ can belong to different \textit{thing} classes, as incorrect semantic classification is penalized only in OccSQ. Similarly, false positive associations (FPA) and false negative associations (FNA) can be defined. Then, the OccAQ can be summarized as:

\begin{equation}
\textit{OccAQ} \!= \!\frac{1}{|G|}\!\!\sum_{g \in G}\!\frac{1}{|gt_{id}(g)|}\!\!\!\!\sum_{\substack{p \in P \\ p \cap g \neq 0}}\!\!\!\textit{TPA}(p, g)\!\cdot\!\text{IoU}(p, g).
\end{equation}

Finally, the OccSTQ is defined as the geometric mean of OccSQ and OccAQ:
\begin{equation}
OccSTQ = \sqrt{OccSQ \times OccAQ}.
\end{equation}

According to the definition of Occ3D~\cite{tian2023occ3d}, we only calculate the metric within the visible region. 
Moreover, OccSTQ can be extended to evaluate scene completion capability without visible region constriction.

\begin{figure*}[thpb]
  \centering
  \includegraphics[scale=0.95]{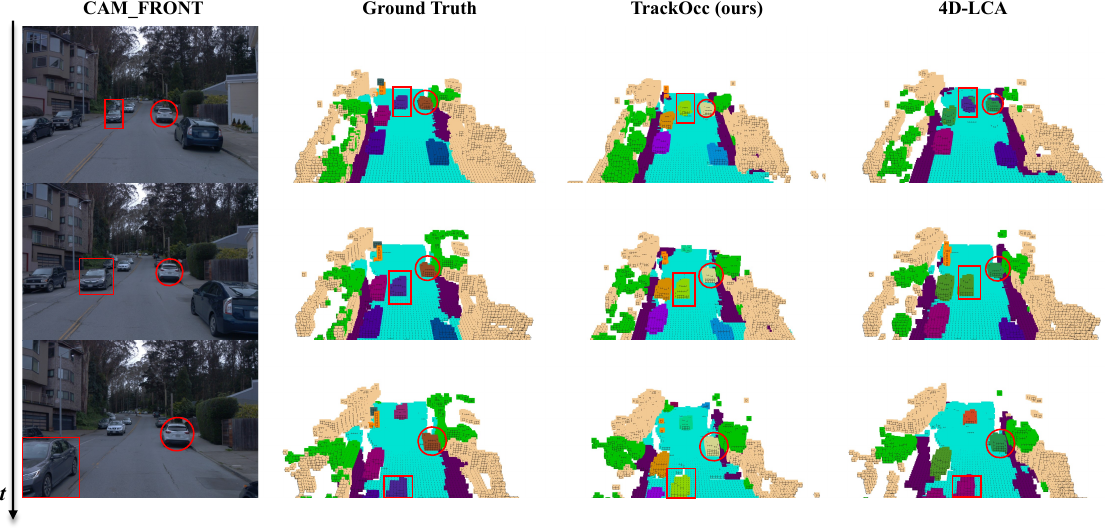}  
    \caption{\textbf{Qualitative results of our method and 4D-LCA.} The same color indicates identical instances across different time steps. Our method effectively tracks both moving and stationary vehicles, while the 4D-LCA method encounters difficulties with moving vehicles. For clarity, only images from the front camera are displayed, although both methods process data from surrounding cameras.}
  \label{figurelabel_FinalV}
  \vspace{-4mm}
\end{figure*}

\subsection{Baselines}

Given the novelty of our proposed task, existing works do not fully align with our objectives. To facilitate a fair comparison, we constructed a series of baselines and demonstrated our method's superior performance. Specifically, we first adopt COTR~\cite{ma2023cotr} as our baseline foundational 3D panoptic occupancy prediction model, since COTR shows better performance than SparseOcc~\cite{yan2021sparse} in Tab.~\ref{table_3DPSCMetricCompare}. Then we adapt it with mainstream temporal association techniques to achieve 4D panoptic occupancy tracking.

 To select the proper 3D panoptic occupancy prediction model, we test the MaskFormer-based methods, COTR~\cite{ma2023cotr} and SparseOcc~\cite{liu2023fully}, using the panoptic quality (PQ) metric~\cite{kirillov2019panoptic} and PQ$^\ast$, as shown in Tab.~\ref{table_3DPSCMetricCompare}. Since PQ's matching requirement, $\text{IoU}>0.5$, is too strict for camera-based methods, resulting in the difficulty of comparing performance on small objects, we transform the matching problem in PQ to maximum weighted bipartite matching~\cite{west2001introduction} and denote the metric as PQ$^\ast$, intend to achieve a fairer comparison.

After the selection, we extend COTR~\cite{ma2023cotr} to the temporal domain with adapted temporal association methods from related fields for fair comparison. 

\begin{itemize}
    \item From video instance segmentation, we directly adopt the bipartite matching approach for cross-frame query association of MinVIS~\cite{huang2022minvis}; 

    \item We use the contrastive training strategy between cross-frame query from CTVIS~\cite{ying2023ctvis} to enhance the query association ability;

    \item To associate instances across frames, an intuitive method is to convert predicted voxels with the same instance ID into bounding boxes and track them using a bounding box tracker, thus, AB3DMOT~\cite{Weng2020_AB3DMOT} is adapted;
 
    \item Drawing from the two-stage association approach of 4D LiDAR panoptic segmentation~\cite{Ayg2024DPLS}, the 4D LiDAR cross-volume association (4D-LCA) is incorporated.

\end{itemize}

\subsection{Main results}
As shown in Table~\ref{table_4DCPOccBaselines}, we present a quantitative comparison between our method and the baselines. TrackOcc outperforms all competitors, particularly in tracking quality. Compared to 4D-LCA~\cite{Ayg2024DPLS}, our method achieves a significant improvement of 3.8 OccSTQ and a 50\% margin in OccAQ, highlighting its strong tracking ability.

Several insights can be drawn from the results. Although CTVIS~\cite{ying2023ctvis} incorporates contrastive loss to enhance instance association across frames, leading to improved performance over MinVIS~\cite{huang2022minvis}, these query-based association methods underperform compared to position-based approaches like AB3DMOT~\cite{Weng2020_AB3DMOT} and 4D-PLS~\cite{Ayg2024DPLS}. This observation highlights the importance of emphasizing localization in our task. It also demonstrates that our method effectively perceives both appearance and location information, significantly improving segmentation and tracking performance.

Moreover, our model achieves a processing speed of 7.5 FPS (inferencing on a stream of around 40 images) on the Waymo dataset with a single RTX 3090.

\subsection{Ablation Studies}

\textbf{Localization-aware loss.} We introduce the localization-aware loss, which adds position supervision for the reference points of 4D panoptic queries, and observe an increase of 3.5 in OccSTQ and more than a 40\% improvement margin in OccAQ, as shown in Tab.~\ref{table_AblationStudies}. This demonstrates the loss benefits 4D panoptic queries by capturing more positionally accurate features, leading to more reliable tracking.

\textbf{The number of training frames.} 
Since increasing the number of training frames better simulates real-world long-sequence driving scenarios, we examine the performance with 3 and 5 training frames, as reported in Tab.~\ref{table_AblationStudies}. The results demonstrate a positive impact on performance when more training frames are added, indicating the scalability of our method in the temporal dimension.

\subsection{Qualitative Comparison}

In Fig.~\ref{figurelabel_FinalV}, we provide a qualitative comparison between our method and 4D-LCA~\cite{Ayg2024DPLS}. TrackOcc demonstrates superior performance in tracking both moving and stationary vehicles. In contrast, 4D-LCA struggles with moving vehicles, failing to maintain consistent instance associations over time, as indicated by the red bounding boxes. These results highlight the strong tracking capability of our method.

\renewcommand{\topfraction}{0.95} 
\renewcommand{\textfraction}{0.05} 
\renewcommand{\floatpagefraction}{0.9} 

\section{CONCLUSIONS}
In this paper, we introduce a new task, \ourtask, which jointly tackles occupancy panoptic segmentation and object tracking from camera input. Besides, We propose \ourwork, a novel camera-based method that achieves 4D panoptic occupancy tracking in a streaming, end-to-end fashion using 4D panoptic queries and localization-aware loss. Experiments demonstrate that \ourwork outperforms other methods. We hope this work will inspire further research in vision-based perception, like end-to-end autonomous driving systems.

\section*{ACKNOWLEDGMENT}
This work is supported by the National Key R\&D Program of China (2022ZD0161700) and Tsinghua University Initiative Scientific Research Program.

\clearpage
\bibliographystyle{IEEEtran}
\bibliography{References}

\begin{thebibliography}{10}
\providecommand{\url}[1]{#1}
\csname url@rmstyle\endcsname
\providecommand{\newblock}{\relax}
\providecommand{\bibinfo}[2]{#2}
\providecommand\BIBentrySTDinterwordspacing{\spaceskip=0pt\relax}
\providecommand\BIBentryALTinterwordstretchfactor{4}
\providecommand\BIBentryALTinterwordspacing{\spaceskip=\fontdimen2\font plus
\BIBentryALTinterwordstretchfactor\fontdimen3\font minus \fontdimen4\font\relax}
\providecommand\BIBforeignlanguage[2]{{%
\expandafter\ifx\csname l@#1\endcsname\relax
\typeout{** WARNING: IEEEtran.bst: No hyphenation pattern has been}%
\typeout{** loaded for the language `#1'. Using the pattern for}%
\typeout{** the default language instead.}%
\else
\language=\csname l@#1\endcsname
\fi
#2}}

\bibitem{li2024DQTrack}
Y.~Li, Z.~Yu, J.~Philion, A.~Anandkumar, S.~Fidler, J.~Jia, and J.~Alvarez, ``End-to-end 3d tracking with decoupled queries,'' in \emph{Proceedings of the IEEE/CVF International Conference on Computer Vision}, 2023, pp. 18\,302--18\,311.

\bibitem{zhangMUTR3DMulticameraTracking2022}
T.~Zhang, X.~Chen, Y.~Wang, Y.~Wang, and H.~Zhao, ``{{MUTR3D}}: {{A Multi-camera Tracking Framework}} via {{3D-to-2D Queries}},'' in \emph{2022 {{IEEE}}/{{CVF Conference}} on {{Computer Vision}} and {{Pattern Recognition Workshops}} ({{CVPRW}})}.\hskip 1em plus 0.5em minus 0.4em\relax New Orleans, LA, USA: IEEE, June 2022, pp. 4536--4545.

\bibitem{ma2023cotr}
Q.~Ma, X.~Tan, Y.~Qu, L.~Ma, Z.~Zhang, and Y.~Xie, ``Cotr: Compact occupancy transformer for vision-based 3d occupancy prediction,'' \emph{arXiv preprint arXiv:2312.01919}, 2023.

\bibitem{cao2022monoscene}
A.-Q. Cao and R.~de~Charette, ``Monoscene: Monocular 3d semantic scene completion,'' in \emph{CVPR}, 2022.

\bibitem{huang2023tri}
Y.~Huang, W.~Zheng, Y.~Zhang, J.~Zhou, and J.~Lu, ``Tri-perspective view for vision-based 3d semantic occupancy prediction,'' \emph{arXiv preprint arXiv:2302.07817}, 2023.

\bibitem{liu2023fully}
H.~Liu, H.~Wang, Y.~Chen, Z.~Yang, J.~Zeng, L.~Chen, and L.~Wang, ``Fully sparse 3d panoptic occupancy prediction,'' \emph{arXiv preprint arXiv:2312.17118}, 2023.

\bibitem{Ayg2024DPLS}
M.~Aygun, A.~Osep, M.~Weber, M.~Maximov, C.~Stachniss, J.~Behley, and L.~Leal-Taix{\'e}, ``4d panoptic lidar segmentation,'' in \emph{Proceedings of the IEEE/CVF Conference on Computer Vision and Pattern Recognition}, 2021, pp. 5527--5537.

\bibitem{yilmaz24mask4former}
K.~Yilmaz, J.~Schult, A.~Nekrasov, and B.~Leibe, ``{Mask4Former: Mask Transformer for 4D Panoptic Segmentation},'' in \emph{{International Conference on Robotics and Automation (ICRA)}}, 2024.

\bibitem{kreuzberg2022stop}
L.~Kreuzberg, I.~E. Zulfikar, S.~Mahadevan, F.~Engelmann, and B.~Leibe, ``4d-stop: Panoptic segmentation of 4d lidar using spatio-temporal object proposal generation and aggregation,'' in \emph{European Conference on Computer Vision Workshop}, 2022.

\bibitem{tian2023occ3d}
X.~Tian, T.~Jiang, L.~Yun, Y.~Wang, Y.~Wang, and H.~Zhao, ``Occ3d: A large-scale 3d occupancy prediction benchmark for autonomous driving,'' \emph{arXiv preprint arXiv:2304.14365}, 2023.

\bibitem{li2023fbocc}
Z.~Li, Z.~Yu, D.~Austin, M.~Fang, S.~Lan, J.~Kautz, and J.~M. Alvarez, ``{FB-OCC}: {3D} occupancy prediction based on forward-backward view transformation,'' \emph{arXiv:2307.01492}, 2023.

\bibitem{cao2024pasco}
A.-Q. Cao, A.~Dai, and R.~de~Charette, ``Pasco: Urban 3d panoptic scene completion with uncertainty awareness,'' in \emph{CVPR}, 2024.

\bibitem{kim2020VPS}
D.~Kim, S.~Woo, J.-Y. Lee, and I.~S. Kweon, ``Video panoptic segmentation,'' in \emph{Proceedings of the IEEE/CVF Conference on Computer Vision and Pattern Recognition}, 2020, pp. 9859--9868.

\bibitem{voigtlaender2020siam}
P.~Voigtlaender, J.~Luiten, P.~H. Torr, and B.~Leibe, ``Siam r-cnn: Visual tracking by re-detection,'' in \emph{Proceedings of the IEEE/CVF conference on computer vision and pattern recognition}, 2020, pp. 6578--6588.

\bibitem{huang2022minvis}
D.-A. Huang, Z.~Yu, and A.~Anandkumar, ``Minvis: A minimal video instance segmentation framework without video-based training,'' \emph{Advances in Neural Information Processing Systems}, vol.~35, pp. 31\,265--31\,277, 2022.

\bibitem{zhang2023dvis}
T.~Zhang, X.~Tian, Y.~Wu, S.~Ji, X.~Wang, Y.~Zhang, and P.~Wan, ``Dvis: Decoupled video instance segmentation framework,'' in \emph{Proceedings of the IEEE/CVF International Conference on Computer Vision}, 2023, pp. 1282--1291.

\bibitem{ying2023ctvis}
K.~Ying, Q.~Zhong, W.~Mao, Z.~Wang, H.~Chen, L.~Y. Wu, Y.~Liu, C.~Fan, Y.~Zhuge, and C.~Shen, ``Ctvis: Consistent training for online video instance segmentation,'' in \emph{Proceedings of the IEEE/CVF International Conference on Computer Vision}, 2023, pp. 899--908.

\bibitem{zhang2023dvis_plus}
T.~Zhang, X.~Tian, Y.~Zhou, S.~Ji, X.~Wang, X.~Tao, Y.~Zhang, P.~Wan, Z.~Wang, and Y.~Wu, ``Dvis++: Improved decoupled framework for universal video segmentation,'' \emph{arXiv preprint arXiv:2312.13305}, 2023.

\bibitem{wang2022detr3d}
Y.~Wang, V.~C. Guizilini, T.~Zhang, Y.~Wang, H.~Zhao, and J.~Solomon, ``Detr3d: 3d object detection from multi-view images via 3d-to-2d queries,'' in \emph{Conference on Robot Learning}.\hskip 1em plus 0.5em minus 0.4em\relax PMLR, 2022, pp. 180--191.

\bibitem{liu2022petr}
Y.~Liu, T.~Wang, X.~Zhang, and J.~Sun, ``Petr: Position embedding transformation for multi-view 3d object detection,'' in \emph{European Conference on Computer Vision}.\hskip 1em plus 0.5em minus 0.4em\relax Springer, 2022, pp. 531--548.

\bibitem{chen2023futr3d}
X.~Chen, T.~Zhang, Y.~Wang, Y.~Wang, and H.~Zhao, ``Futr3d: A unified sensor fusion framework for 3d detection,'' in \emph{proceedings of the IEEE/CVF conference on computer vision and pattern recognition}, 2023, pp. 172--181.

\bibitem{lai2023spherical}
X.~Lai, Y.~Chen, F.~Lu, J.~Liu, and J.~Jia, ``Spherical transformer for lidar-based 3d recognition,'' in \emph{Proceedings of the IEEE/CVF Conference on Computer Vision and Pattern Recognition}, 2023, pp. 17\,545--17\,555.

\bibitem{zhang2022mutr3d}
T.~Zhang, X.~Chen, Y.~Wang, Y.~Wang, and H.~Zhao, ``Mutr3d: A multi-camera tracking framework via 3d-to-2d queries,'' in \emph{Proceedings of the IEEE/CVF Conference on Computer Vision and Pattern Recognition}, 2022, pp. 4537--4546.

\bibitem{ding2024ada}
S.~Ding, L.~Schneider, M.~Cordts, and J.~Gall, ``Ada-track: End-to-end multi-camera 3d multi-object tracking with alternating detection and association,'' in \emph{Proceedings of the IEEE/CVF Conference on Computer Vision and Pattern Recognition}, 2024, pp. 15\,184--15\,194.

\bibitem{marcuzzi2023ral-meem}
\BIBentryALTinterwordspacing
R.~Marcuzzi, L.~Nunes, L.~Wiesmann, E.~Marks, J.~Behley, and C.~Stachniss, ``{Mask4D: End-to-End Mask-Based 4D Panoptic Segmentation for LiDAR Sequences},'' \emph{IEEE Robotics and Automation Letters}, vol.~8, no.~11, pp. 7487--7494, 2023. [Online]. Available: \url{https://www.ipb.uni-bonn.de/wp-content/papercite-data/pdf/marcuzzi2023ral-meem.pdf}
\BIBentrySTDinterwordspacing

\bibitem{zhu20234d}
M.~Zhu, S.~Han, H.~Cai, S.~Borse, M.~G. Jadidi, and F.~Porikli, ``4d panoptic segmentation as invariant and equivariant field prediction,'' in \emph{International Conference on Computer Vision}, 2023.

\bibitem{LSS}
J.~Philion and S.~Fidler, ``Lift, splat, shoot: Encoding images from arbitrary camera rigs by implicitly unprojecting to 3d,'' in \emph{Computer Vision -- ECCV 2020}, A.~Vedaldi, H.~Bischof, T.~Brox, and J.-M. Frahm, Eds.\hskip 1em plus 0.5em minus 0.4em\relax Cham: Springer International Publishing, 2020, pp. 194--210.

\bibitem{li2023bevdepth}
Y.~Li, Z.~Ge, G.~Yu, J.~Yang, Z.~Wang, Y.~Shi, J.~Sun, and Z.~Li, ``Bevdepth: Acquisition of reliable depth for multi-view 3d object detection,'' \emph{Proceedings of the AAAI Conference on Artificial Intelligence}, vol.~37, no.~2, pp. 1477--1485, Jun. 2023.

\bibitem{li2022bevformer}
Z.~Li, W.~Wang, H.~Li, E.~Xie, C.~Sima, T.~Lu, Y.~Qiao, and J.~Dai, ``Bevformer: Learning bird’s-eye-view representation from multi-camera images via spatiotemporal transformers,'' in \emph{European conference on computer vision}.\hskip 1em plus 0.5em minus 0.4em\relax Springer, 2022, pp. 1--18.

\bibitem{vaswani2017attention}
A.~Vaswani, ``Attention is all you need,'' \emph{Advances in Neural Information Processing Systems}, 2017.

\bibitem{deformttn}
Z.~Xia, X.~Pan, S.~Song, L.~E. Li, and G.~Huang, ``Vision transformer with deformable attention,'' in \emph{Proceedings of the IEEE/CVF conference on computer vision and pattern recognition}, 2022, pp. 4794--4803.

\bibitem{cheng2021mask2former}
B.~Cheng, I.~Misra, A.~G. Schwing, A.~Kirillov, and R.~Girdhar, ``Masked-attention mask transformer for universal image segmentation,'' \emph{arXiv}, 2021.

\bibitem{kuhn1955hungarian}
H.~W. Kuhn, ``The hungarian method for the assignment problem,'' \emph{Naval research logistics quarterly}, vol.~2, no. 1-2, pp. 83--97, 1955.

\bibitem{Weng2020_AB3DMOT}
X.~Weng, J.~Wang, D.~Held, and K.~Kitani, ``{3D Multi-Object Tracking: A Baseline and New Evaluation Metrics},'' \emph{IROS}, 2020.

\bibitem{WODsun_2020_CVPR}
P.~Sun, H.~Kretzschmar, X.~Dotiwalla, A.~Chouard, V.~Patnaik, P.~Tsui, J.~Guo, Y.~Zhou, Y.~Chai, B.~Caine, V.~Vasudevan, W.~Han, J.~Ngiam, H.~Zhao, A.~Timofeev, S.~Ettinger, M.~Krivokon, A.~Gao, A.~Joshi, Y.~Zhang, J.~Shlens, Z.~Chen, and D.~Anguelov, ``Scalability in perception for autonomous driving: Waymo open dataset,'' in \emph{Proceedings of the IEEE/CVF Conference on Computer Vision and Pattern Recognition (CVPR)}, June 2020.

\bibitem{He_2016_CVPR}
K.~He, X.~Zhang, S.~Ren, and J.~Sun, ``Deep residual learning for image recognition,'' in \emph{Proceedings of the IEEE Conference on Computer Vision and Pattern Recognition (CVPR)}, June 2016.

\bibitem{loshchilov2019decoupledweightdecayregularization}
\BIBentryALTinterwordspacing
I.~Loshchilov and F.~Hutter, ``Decoupled weight decay regularization,'' 2019. [Online]. Available: \url{https://arxiv.org/abs/1711.05101}
\BIBentrySTDinterwordspacing

\bibitem{hurtado2020mopt}
J.~V. Hurtado, R.~Mohan, W.~Burgard, and A.~Valada, ``Mopt: Multi-object panoptic tracking,'' \emph{arXiv preprint arXiv:2004.08189}, 2020.

\bibitem{weber2021step}
M.~Weber, J.~Xie, M.~Collins, Y.~Zhu, P.~Voigtlaender, H.~Adam, B.~Green, A.~Geiger, B.~Leibe, D.~Cremers, \emph{et~al.}, ``Step: Segmenting and tracking every pixel,'' \emph{arXiv preprint arXiv:2102.11859}, 2021.

\bibitem{yan2021sparse}
X.~Yan, J.~Gao, J.~Li, R.~Zhang, Z.~Li, R.~Huang, and S.~Cui, ``Sparse single sweep lidar point cloud segmentation via learning contextual shape priors from scene completion,'' in \emph{Proceedings of the AAAI Conference on Artificial Intelligence}, vol.~35, no.~4, 2021, pp. 3101--3109.

\bibitem{kirillov2019panoptic}
A.~Kirillov, K.~He, R.~Girshick, C.~Rother, and P.~Doll{\'a}r, ``Panoptic segmentation,'' in \emph{Proceedings of the IEEE/CVF conference on computer vision and pattern recognition}, 2019, pp. 9404--9413.

\bibitem{west2001introduction}
D.~B. West, ``Introduction to graph theory,'' 2001.

\end{thebibliography}
\addtolength{\textheight}{-12cm}   

\end{document}